\documentclass[Afour,sageh,times]{sagej}

\usepackage{amsmath, amssymb}
\usepackage{graphicx}
\usepackage{algorithm}
\usepackage{algpseudocode}
\usepackage{textcomp}
\usepackage{bm}
\usepackage{booktabs}
\usepackage{array}
\usepackage{subcaption}
\usepackage[colorlinks,bookmarksopen,bookmarksnumbered,citecolor=red,urlcolor=red]{hyperref}

\newcommand{\Transp}{\mathsf{T}}

\def\BibTeX{{\rm B\kern-.05em{\sc i\kern-.025em b}\kern-.08em
    T\kern-.1667em\lower.7ex\hbox{E}\kern-.125emX}}

\begin{document}

\runninghead{Edridge and Kok: SL(C)AMma}

\title{SL(C)AMma: Simultaneous Localisation, (Calibration) and Mapping With a Magnetometer Array}

\author{Thomas Edridge\affilnum{1} and Manon Kok\affilnum{1}}

\affiliation{\affilnum{1}DCSC, Delft University of Technology, the Netherlands.}

\corrauth{Thomas Edridge. Email: \href{mailto:t.i.edridge@tudelft.nl}{\textcolor{black}{t.i.edridge@tudelft.nl}}}

\begin{abstract}
Indoor localisation techniques suffer from attenuated Global Navigation Satellite System (GNSS) signals and from the accumulation of unbounded drift by integration of proprioceptive sensors. Magnetic field-based Simultaneous Localisation and Mapping (SLAM) reduces drift through loop closures by revisiting previously seen locations, but extended exploration of unseen areas remains challenging. Recently, magnetometer arrays have demonstrated significant benefits over single magnetometers, as they can directly estimate the odometry. However, inconsistencies between magnetometer measurements negatively affect odometry estimates and complicate loop closure detection. We propose two filtering algorithms: The first focuses on magnetic field-based SLAM using a magnetometer array (SLAMma). The second extends this to jointly estimate the magnetometer calibration parameters (SLCAMma). We demonstrate, using Monte Carlo simulations, that the calibration parameters can be accurately estimated when there is sufficient orientation excitation, and that magnetometers achieve inter-sensor measurement consistency regardless of the type of motion. Experimental validation on ten datasets confirms these results, and we demonstrate that in cases where single magnetometer SLAM fails, SLAMma and SLCAMma provide good trajectory estimates with, more than 80\% drift reduction compared to integration of proprioceptive sensors.
\end{abstract}


\maketitle

\section{Introduction}
\label{sec:introduction}
\begin{figure}[t!]
    \centering
    \includegraphics[width=\linewidth]{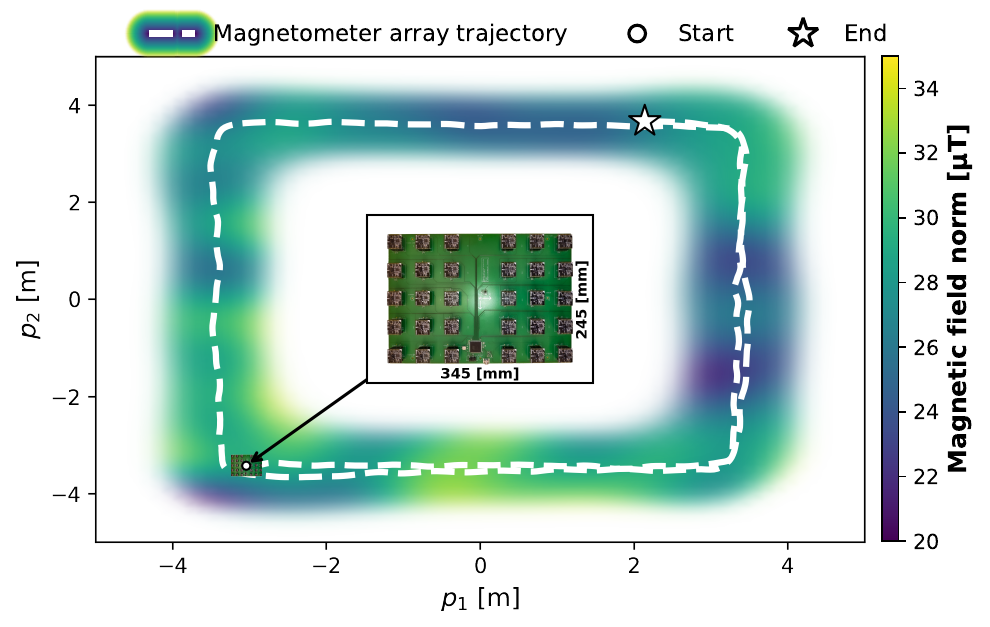}
    \caption{Example of a 30-magnetometer array moving in a square in an indoor environment with a spatially disturbed magnetic field. The background colour corresponds to the norm of the magnetic field, and the transparency is proportional to the uncertainty of the magnetic field predictions.}
    \label{fig:magArrayIntro}
\end{figure}

Indoor localisation is important for applications such as autonomous navigation of mobile agents, for example, in multi-agent robotic logistics in warehouses \citep{krnjaic2024scalable}, or firefighters responding to emergencies \citep{nilsson2014accurate}. Building structures attenuate the Global Navigation Satellite System (GNSS) signals, rendering them unreliable or unavailable indoors \citep{storms2010magnetic}. Nowadays, most smart devices are equipped with an Inertial Measurement Unit (IMU) \citep{kok2017using}, consisting of an accelerometer and a gyroscope. Mobile robots are additionally equipped with wheel encoders \citep{huang2023indoor}. These proprioceptive sensors provide position and orientation information, but this is widely known to drift over time~\citep{kok2017using}. Simultaneous Localisation and Mapping (SLAM) offers a promising framework to reduce the amount of drift \citep{cadena2017past}, by utilising proprioceptive sensors in combination with various exteroceptive sensors, such as LiDARs, cameras, and ultrawideband (UWB). However, these sensors each have their drawbacks, which we overcome using an array of magnetometers, see also Figure~\ref{fig:magArrayIntro}. Unlike LiDARs, which are often bulky and expensive \citep{lin2022r, hess2016real, pan2021mulls, shan2021lvi}, cameras, that perform poorly in visually feature-poor/low-light environments \citep{shan2021lvi,chung2022orbeez,yan2022joint,xu2020decentralized}, or UWB sensors, which require an infrastructure with a fixed anchor \citep{xu2020decentralized, yang2021uvip, cao2020accurate, nguyen2020tightly}, the magnetometer is low-cost, compact, and does not require line of sight.

The magnetometer, which measures a 3D vector of the ambient magnetic field, has recently shown promise for indoor localisation \citep{siebler2022bayesian, dammann2024cramer, li2012feasible, angermann2012characterization, solin2018modeling, viset2022extended}. While magnetometers are traditionally used as a compass, these recent methods create a map of the disturbances in the magnetic field of the indoor environment, which can be used for indoor localisation. Ferromagnetic materials cause these disturbances and are generally temporally stable \citep{ouyang2022survey}, except for disruptive and moving sources such as elevators. An example of an indoor magnetic field map is illustrated in Figure~\ref{fig:magArrayIntro}.

Recently, it has been demonstrated that a key advantage of a magnetometer array over a single magnetometer is its ability to extract odometry information through spatial variations in the ambient magnetic field \citep{skog2018magnetic, skog2021magnetic}. The idea is that spatially distributed magnetometers record two consecutive ``images'' of the magnetic field, one before and one after displacement. By comparing the two images, the change in position and orientation can be estimated. This has great potential for robotics as their platforms can easily carry multiple, low-cost magnetometers. This work extends the single magnetometer SLAM algorithm of \citet{viset2022extended}, which represents the magnetic field map using a reduced rank Gaussian process (GP) formulation from \citet{kok2018scalable}, by extending it to allow for the inclusion of measurements from a magnetometer array.

A challenge with using multiple magnetometers in magnetic field-based SLAM is inter-sensor measurement consistency \citep{ouyang2022survey}, which requires all magnetometers to measure the same magnetic field values when placed in the same location. This is important for two reasons. Firstly, uncalibrated magnetometers negatively affect the exploration phase of the magnetometer array, increasing the error accumulation in the estimated odometry~\citep{edridge2025position}. Secondly, it complicates loop closure detection. To be more specific, if the magnetometer array revisits a previous location in a different orientation with uncalibrated magnetometers, a loop closure is unlikely to be detected. We therefore extend our magnetometer array filtering algorithm by incorporating online magnetometer calibration of scaling parameters and biases. We validate both algorithms using multiple experimental datasets. The python implementation used for simulation and experimental validation is made publicly available \footnote{Available online at \href{https://github.com/Tedridge/SL-C-AMma}{https://github.com/Tedridge/SL-C-AMma}}.

\section{Related work}
Magnetometer arrays were first used to estimate the velocity to aid indoor localisation \citep{vissiere2007using, dorveaux2011presentation, dorveaux2011combining, chesneau2016motion, zmitri2021bilstm, zmitri2020magnetic}. Inspired by this approach, \citet{skog2018magnetic, skog2021magnetic} show that the odometry (change in position and orientation) can be estimated directly. Following up prior work, \citet{edridge2025position} analysed the precision of the odometry estimates. The work in \citet{huang2024mains} combines a magnetometer array with an inertial navigation system. Furthermore, \citet{skog2025connection} explores a theoretical comparison between magnetic-field odometry and magnetic-field SLAM. Finally, \citet{huang2025inertial} uses a magnetometer array to estimate the odometry explicitly and fuses the estimates in a SLAM framework. In contrast, we estimate the pose change implicitly by maximising the posterior and present a closed-form algorithm for magnetic-field SLAM using a magnetometer array. Furthermore, we extend our algorithm to include online magnetometer calibration.

Existing magnetometer calibration methods, such as those using ellipsoid fitting \citep{renaudin2010complete} or maximum likelihood estimation \citep{kok2016magnetometer}, require a disturbance-free magnetic field, necessitating a dataset away from the desired indoor localisation area. Alternatively, the calibration parameters can be estimated online in a known magnetic field map \citep{siebler2020joint, siebler2021evaluation, siebler2023magnetic}. The work in \citet{vallivaara2024saying} shows it is possible to estimate the magnetometer bias online in a magnetic field SLAM framework. Magnetometer array calibration is addressed in \citet{wang2017fast, mu2018novel, mu2019calibration, chesneau2019calibration, huang2025joint}. In particular, \citet{chesneau2019calibration} demonstrates that the position, the magnetic field map, and all calibration parameters are identifiable up to an arbitrary scaling factor of the magnetic field strength. However, they assume access to the true orientations of the magnetometer array. This is often unrealistic in practice, as position and orientation estimates are obtained through inertial sensors or wheel encoders, and these estimates, therefore, suffer from integration drift. Recently, \citet{lyu2026roslac} addressed this limitation by accumulating noisy wheel odometry to jointly localise and calibrate a magnetometer array, though their method requires a known, pre-built magnetic field map. In contrast, our approach simultaneously estimates the magnetic field map online. By assuming access to noisy changes in position and orientation, we solve the non-linear SLAM and calibration problem directly within a filtering framework.

\section{Problem formulation}
\label{sec:main}
\begin{figure}[h!]
    \centering\includegraphics[width=\linewidth]{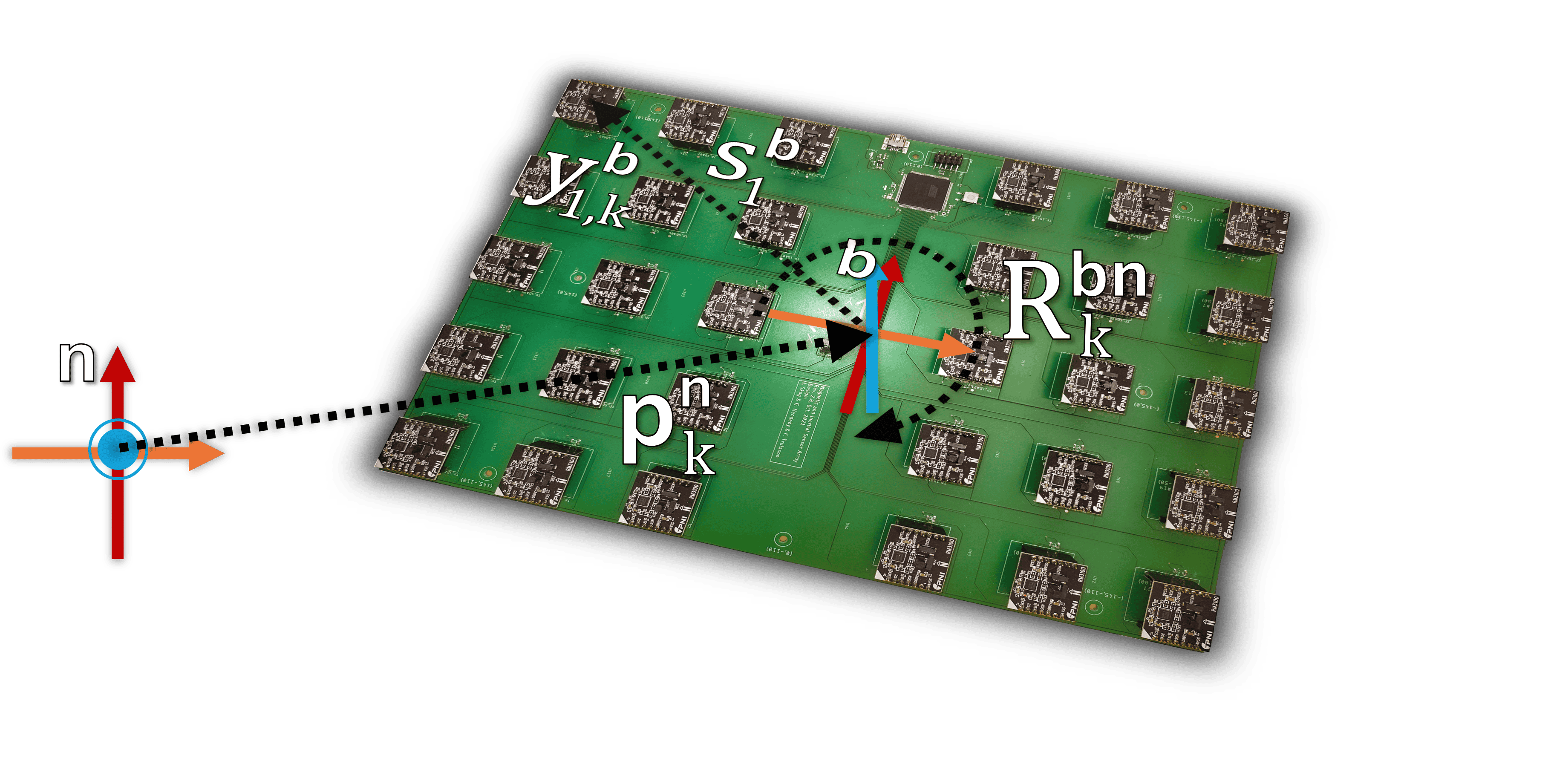}
    \caption{Magnetometer array of size $345\mathrm{mm} \times 245\mathrm{mm} $, consisting of 30 magnetometers. The body frame is defined in the centre of the sensor board. We illustrate the measurement $\mathbf{y}_{1,k}^{\text{b}}$ and position $\mathbf{s}_{1}^{\text{b}}$ of the first magnetometer.}
    \label{fig:rigidBody}
\end{figure}

To solve the problem of magnetic-field SLAM using a magnetometer array, we maximise the posterior probability density function (pdf) 
\begin{equation}
    p\left(\mathbf{x}_k  |\mathbf{y}^{\text{b}}_{1:k}, \Delta \mathbf{p}^{\text{b}}_{1:k-1}, \Delta \mathbf{R}_{1:k-1}\right),
    \label{eq:posteriorPDF}
\end{equation}
where $p(a \mid b)$ denotes the conditional pdf of $a$ given $b$ and for notational brevity we define $\mathbf{y}^{\text{b}}_{1:k} = \{ \{\mathbf{y}^{\text{b}}_{i,k} \}_{i=1}^{N_\text{mag}} \}_{k=1}^{N_k} $, with ${\mathbf{y}^{\text{b}}_{i,k}\in \mathbb{R}^{3 \times 1} }$, the measurement set of all $N_{\text{mag}}$ magnetometers on the array for all time-steps $N_{K}$. We use superscript b to denote the body frame, of which the origin and axes are aligned with the origin and axes of the array as shown in Figure~\ref{fig:rigidBody}. Furthermore, we assume to have access to the change in position $\Delta \mathbf{p}^{\text{b}}_{1:k-1}$ and change in orientation $\Delta \mathbf{R}_{1:k-1}$ of the magnetometer array. Moreover, we define the state $\mathbf{x}_k$ in~\eqref{eq:posteriorPDF} as
\begin{equation}
    \mathbf{x}_k = \left\{ \mathbf{p}^{\text{n}}_k, \mathbf{R}^{\text{bn}}_k, \mathbf{m}, \mathbf{d}^{\text{b}}, \mathbf{b}^{\text{b}}\right\},
    \label{eq:stateFull}
\end{equation}
where $\mathbf{p}^{\text{n}}_k$ is the location of the magnetometer array at time-step $k$, with superscript n denoting the navigation frame as depicted in Figure~\ref{fig:rigidBody}. Furthermore, $\mathbf{R}^{\text{bn}}_k \in \mathrm{SO}(3)$ is the rotation matrix, from navigation frame to body frame at time-step $k$. The vector $\mathbf{m}\in \mathbb{R}^{N_{\text{weights}} \times 1}$ describes the weights used to represent the ambient magnetic field map. We use the shorthand notations $\mathbf{d}^{\text{b}} \triangleq  \{ \mathbf{d}^{\text{b}}_i \}_{i=1}^{N_\text{mag}}$ and $\mathbf{b}^{\text{b}} \triangleq \{ \mathbf{b}^{\text{b}}_i \}_{i=1}^{N_\text{mag}}$, with $\mathbf{d}^{\text{b}}_i, \mathbf{b}^{\text{b}}_i \in \mathbb{R}^{3\times1}$ to denote the scaling parameters and biases of magnetometer $i \in \{1, \hdots, N_{\text{mag}} \}$ on the array. The scaling parameters compensate for effects induced by soft iron materials, and the bias compensates for hard iron effects, disturbing the magnetometer measurements.

The relative positions of the magnetometers $\mathbf{s}_i^{\text{b}}$  are assumed to be known in the body frame as shown in Figure~\ref{fig:rigidBody}. Hence, the position of a magnetometer on the magnetometer array at time step $k$ can be described in the navigation frame by $\mathbf{p}^{\text{n}}_k + \mathbf{R}^{\text{nb}}_k\mathbf{s}_i^{\text{b}}$. We can now write the dynamic pdf in terms of the changes in positions $\Delta \mathbf{p}^{\text{b}}_k$ and change in orientations $\Delta \mathbf{R}_k$ as
\begin{equation}
    p(\mathbf{p}^{\text{n}}_{k+1},\mathbf{R}^{\text{bn}}_{k+1} |\mathbf{p}^{\text{n}}_{k},\mathbf{R}^{\text{bn}}_k, \Delta \mathbf{p}^{\text{b}}_{k}, \Delta \mathbf{R}_{k}).
\end{equation}
We can also write the pdf of our measurements as
\begin{equation}
    p(\mathbf{y}^{\text{b}}_{k} |  \mathbf{x}_k) = \prod_{i = 1}^{N_{\text{mag}}}  p(\mathbf{y}^{\text{b}}_{i,k} |  \mathbf{x}_k),
    \label{eq:measPDF}
\end{equation}
where we use the assumption that the measurements of the magnetometers on the array are independent.

We present two algorithms of magnetic field SLAM with a magnetometer array: SLAMma, which assumes pre-calibrated magnetometers and SLCAMma, which calibrates the magnetometers online. We want to highlight that, whenever we use the term loop-closure, this is not an explicit loop-closure as is common practice in many robotics applications. Contrarily, the loop-closure is done implicitly by maximising the posterior from~\eqref{eq:posteriorPDF} when re-visiting a previously seen location. Additionally, we highlight that our SL(C)AMma algorithm does not explicitly compute magnetic field odometry, as for instance in~\cite{huang2025inertial}. However, this is also done implicitly by maximising the posterior from~\eqref{eq:posteriorPDF}.

\section{Modelling}
\label{sec:models}
\subsection{Dynamic model}
\label{sec:dynamicModel}
We assume a dynamic model where we have noisy measurements of changes in position and orientation at every time step as
\begin{equation}
\begin{aligned}
    \mathbf{p}^{\text{n}}_{k+1} &= \mathbf{p}^{\text{n}}_{k} +\mathbf{R}_{k}^{\text{nb}} \left( \Delta \mathbf{p}^{\text{b}}_{k} + \bm{\epsilon}^{\text{b}}_{k} \right) , 
    &\quad \mathbf{\epsilon}^{\text{b}}_{k} &\sim \mathcal{N}\left(0, \bm{\Sigma}_{\text{pos}}\right), \\
    \mathbf{R}_{k+1}^{\text{bn}} &= \text{exp}_{\mathbf{R}}(\bm{\nu}^\text{b}_{k}) \Delta \mathbf{R}_{k} \mathbf{R}_{k}^{\text{bn}}, 
    &\quad \bm{\nu}^\text{b}_{k} &\sim \mathcal{N}\left(0, \bm{\Sigma}_{\text{rot}}\right),
\end{aligned}
\label{eq:dynamicModel}
\end{equation}
where $\mathcal{N}(\cdot,\cdot)$ denotes the Gaussian distribution and where the operator $\text{exp}_{\mathbf{R}}(\cdot)$ as defined in \cite{kok2017using} maps an orientation deviation to a rotation matrix in $\mathrm{SO}(3)$. 
Furthermore, we assume that the ambient magnetic field is time-invariant \citep{ouyang2022survey} as are the calibration parameters, i.e., the dynamic model of these parameters is: ${\mathbf{m}_{k+1} = \mathbf{m}_{k}}$, ${
    \mathbf{d}^{\text{b}}_{i,k+1} = \mathbf{d}^{\text{b}}_{i,k}}$, and 
    ${\mathbf{b}^{\text{b}}_{i,k+1} = \mathbf{b}^{\text{b}}_{i,k} }$.  

\subsection{Magnetic field model}
\label{sec:magneticFieldModel}
Consistent with \cite{solin2018modeling}, the ambient magnetic field is modeled as the gradient of a scalar potential $\nabla \bm{\phi}(\mathbf{p})$, using a GP. Modelling the magnetic field as the gradient of a scalar potential ensures a curl-free kernel assumption, which satisfies the Maxwell equations, and, as shown by \cite{solin2018modeling}, it is effective for modelling the magnetic field as a curl-free vector field. The GP prior has a distribution
\begin{equation}
    \bm{\phi}(\mathbf{p}) \sim \mathcal{GP} \left(0, \kappa_{\text{lin}}(\mathbf{p},\mathbf{p}') + \kappa_\text{se}(\mathbf{p}, \mathbf{p}') \right).
    \label{eq:GPmodelSE}
\end{equation}
Where, $\mathcal{GP}(\cdot,\cdot)$ denotes the GP distribution. Furthermore, we have kernel ${\kappa_{\text{lin}}(\mathbf{p},\mathbf{p}') = \sigma^2_{\text{lin}} \mathbf{p}^\Transp \mathbf{p}'}$ and the squared exponential kernel is defined as
\begin{equation}
\begin{aligned}
    \kappa_{\text{se}}(\mathbf{p}, \mathbf{p}') =& \sigma^2_\text{se} \text{exp}\left( \frac{||\mathbf{p} - \mathbf{p}'||^2}{-2\ell^2} \right).
    \label{eq:squaredExponential}
\end{aligned}
\end{equation}
which models disturbances caused by ferromagnetic materials present in the structure of buildings. The kernel's GP hyper-parameters are the length-scale $\ell$ and the signal variance $\sigma_\text{se}^2$. As we are interested in SLAM (and calibration) using a magnetometer array of 30 magnetometers, each collecting magnetic field measurements at high frequencies, we approximate the GP in a reduced rank form from \cite{solin2018modeling} to reduce the computational complexity and memory requirements. This reduced rank form is based on a Laplacian eigenvalue decomposition, where the basis functions stretch over a known finite cubical domain ${\bm{\Omega} = [L_{l,1},L_{u,1}]\times [L_{l,2},L_{u,2}] \times [L_{l,3},L_{u,3}] }$. The weights have prior distribution $\mathbf{m} \sim \mathcal{N}(0, \bm{\Lambda}_{\text{w}})$ with
\begin{equation}
     \bm{\Lambda}_{\text{w}} = \text{diags}\left(\sigma^2_{\text{lin}}, \sigma^2_{\text{lin}}, \sigma^2_{\text{lin}}, S_{\text{se}}(\sqrt{\lambda_1}), \hdots, S_{\text{se}}(\sqrt{\lambda_m}) \right)
     \label{eq:priorWeights}
\end{equation}
with the spectral density $S_{\text{se}}(\hdots)$ and $\lambda_j$ the $j_\text{th}$ eigenvalue both defined in \cite{solin2018modeling}. Here, the operator $\text{diags}(\cdot)$ places the elements on the diagonal of a square matrix. By truncating for the highest $N_{\text{weights}}$ eigenvalues, we approximate the GP in reduced rank form as ${\nabla\bm{\phi}(\cdot) \approx \nabla\bm{\Phi}(\cdot)\mathbf{m}}$. Here, $\nabla\bm{\Phi}(\cdot)$ is defined in \cite{solin2018modeling} and the weights $\mathbf{m}$ are part of the state from~\eqref{eq:stateFull}.

\subsection{Magnetometer array measurement model}
\label{sec:measurementModel}
Let us assume that all $N_{\text{mag}}$ magnetometers measure the magnetic field synchronously at each time step $k$, where each measurement is independent according to~\eqref{eq:measPDF}, the measurement vector of all magnetometers on the array is then ${\mathbf{y}_{k}^{\text{b}} \triangleq \big[( \mathbf{y}_{1,k}^{\text{b}})^\Transp \;\cdots\; (\mathbf{y}_{N_{\text{mag}},k}^{\text{b}})^\Transp \big]^\Transp}$. Let us also assume we know matrix ${\mathbf{S}^{\text{b}} \triangleq \begin{bmatrix} \mathbf{s}_{1}^{\text{b}} & \hdots & \mathbf{s}_{N_{\text{mag}}}^{\text{b}} \end{bmatrix}}$, which describes the relative locations of all magnetometers on the array in the body frame and let us define the matrix ${\mathbf{P}_{k}^{\text{n}} = \mathbf{p}_{k}^{\text{n}} \otimes \mathbf{1}_{N_{\text{mag}}}^\Transp}$. Here, $\mathbf{p}_{k}^{\text{n}} $ is the magnetometer array centre location from~\eqref{eq:stateFull} and the notation $\mathbf{1}_{N_{\text{mag}}}$ is a vector of ones of length $N_{\text{mag}}$. 
This allows us to define the measurement model of the magnetometer array as a reduced rank GP in terms of our state from~\eqref{eq:stateFull} as
\begin{equation}
\begin{aligned}
        \mathbf{y}_{k}^{\text{b}} &\approx
      \mathbf{D}^{\text{b}} \left( \mathbf{R}_{k}^{\text{bn}} \nabla \bm{\Phi}\left(\mathbf{P}_{k}^{\text{n}} + \mathbf{R}_k^{\text{nb}} \mathbf{S}^{\text{b}}\right) \mathbf{m} + \mathbf{b}^{\text{b}} \right) +  \mathbf{e}^{\text{b}}_{k}, \\
    \mathbf{e}^{\text{b}}_{k} &\sim \mathcal{N}(0, \sigma^2_{\text{y}} \mathbf{I}_{3N_\text{mag}}),
    \end{aligned}
    \label{eq:magMeas}
\end{equation}
with assumed Gaussian measurement noise $\mathbf{e}^{\text{b}}_{k}$ and ${\mathbf{R}_{k}^{\text{nb}} \triangleq (\mathbf{R}_{k}^{\text{bn}})^\Transp}$. The calibration parameters $\mathbf{b}^{\text{b}}$ and $\mathbf{d}^{\text{b}}$ are defined as ${\mathbf{b}^{\text{b}} = \big[( \mathbf{b}_{1}^{\text{b}})^\Transp \;\cdots\; (\mathbf{b}_{N_{\text{mag}}}^{\text{b}})^\Transp \big]^\Transp}$ and ${\mathbf{D}^{\text{b}} = \text{diags}(\mathbf{d}_1, \hdots, \mathbf{d}_{N_{\text{mag}}} )}^{-1}$. 

\section{Methods}
\label{sec:S(C)LAMma}
To maximise our posterior from~\eqref{eq:posteriorPDF} using the models described in Section~\ref{sec:models}, we estimate our state $\mathbf{x}_k$ from~\eqref{eq:stateFull} in a magnetic field SLAM filter with a magnetometer array. We achieve this using an iterated version~\citep{bell1993iterated} of the error state extended information filter, similar to~\citet{viset2023distributed}. To this end, we introduce the error state of \eqref{eq:stateFull} as
\begin{equation}
    \delta \mathbf{x}_k = 
    \begin{bmatrix}
    \left( \delta \mathbf{p}^{\text{n}}_k \right)^\Transp, 
    \left(\bm{\eta}^\text{b}_k \right)^\Transp, 
    \left( \delta \mathbf{m}_k \right)^\Transp, 
    \left( \delta \mathbf{d}_k^{\text{b}}\right)^\Transp, 
    \left( \delta \mathbf{b}_k^{\text{b}}\right)^\Transp
    \end{bmatrix}^\Transp.
    \label{eq:deltaState}
\end{equation}
Here, $\delta \mathbf{p}^{\text{n}}_k$, $\bm{\eta}^\text{b}_k$, $\delta \mathbf{m}_k$, $\delta \mathbf{d}_k^{\text{b}}$ and $\delta \mathbf{b}_k^{\text{b}}$ respectively denote the estimation errors of the position, orientation, magnetic field weights, scaling and bias. 

Note that we use the information form to avoid numerical instability that occurs in the standard Kalman filter. As the magnetometer array Jacobians are highly correlated, the innovation covariance becomes ill-conditioned. 

\subsection{Dynamic update}
\label{sec:dynamicUpdate}
Following the approach in \citet{viset2022extended}, the state from~\eqref{eq:stateFull} is updated directly using the dynamic equations from Section~\ref{sec:dynamicModel} as
\begin{subequations}
\label{eq:dynamicUpdates}
\begin{align}
    \hat{\mathbf{p}}^{\text{n}}_{k+1|k}      &= \hat{\mathbf{p}}^{\text{n}}_{k|k} + \hat{\mathbf{R}}^{\text{nb}}_{k|k} \Delta \mathbf{p}^{\text{b}}_k, \label{eq:dynamicUpdatePosition} \\
    \hat{\mathbf{R}}^{\text{bn}}_{k+1|k}     &= \Delta \mathbf{R}_k  \hat{\mathbf{R}}^{\text{bn}}_{k|k}, \label{eq:dynamicUpdateRotation} \\
    \hat{\mathbf{m}}_{k+1|k}                 &= \hat{\mathbf{m}}_{k|k}, \label{eq:dynamicUpdateWeights} \\
    \hat{\mathbf{d}}^{\text{b}}_{k+1|k}                 &= \hat{\mathbf{d}}^{\text{b}}_{k|k}, \label{eq:dynamicUpdateScale} \\
    \hat{\mathbf{b}}^{\text{b}}_{k+1|k}                 &= \hat{\mathbf{b}}^{\text{b}}_{k|k}, \label{eq:dynamicUpdateBias}
\end{align}
\end{subequations}
where the hat operator $\hat{\cdot}$ denotes an estimate. Note that we present two algorithms, one with pre-calibrated magnetometers and the other with online magnetometer calibration. In the former, \eqref{eq:dynamicUpdateScale} and \eqref{eq:dynamicUpdateBias} are omitted. 

The information matrix $\mathcal{I}$ is defined as the inverse of the state covariance matrix from~\eqref{eq:deltaState}, and is updated in the dynamic update as follows 
\begin{equation}
    \bm{\mathcal{I}}_{k+1|k} = \tilde{\bm{\mathcal{I}}}_{k|k} -  \Delta \tilde{\bm{\mathcal{I}}}_{k} ,
    \label{eq:dynamicInformationUpdate}
\end{equation}
with ${\Delta \tilde{\bm{\mathcal{I}}}_{k} \triangleq \tilde{\bm{\mathcal{I}}}_{k|k} \mathbf{G}_k (\mathbf{Q}^{-1}_k + \mathbf{G}^{\Transp}_k  \tilde{\bm{\mathcal{I}}}_{k|k}  \mathbf{G}_k)^{-1} \mathbf{G}^{\Transp}_k \tilde{\bm{\mathcal{I}}}_{k|k}}$ and ${\tilde{\bm{\mathcal{I}}}_{k|k} \triangleq \mathbf{F}_k^{-\Transp}\bm{\mathcal{I}}_{k|k} \mathbf{F}_k^{-1}}$, 
where we used the Woodbury matrix identity~\citep{petersen2008matrix}, following~\citet{thrun2004simultaneous}. The matrices $\mathbf{F}_k$ and $\mathbf{G}_k$ and covariance $\mathbf{Q}$ are defined as
\begin{equation}
\begin{aligned}
   \mathbf{F}_k &= \begin{bmatrix}
        \mathbf{I}_3 & \hat{\mathbf{R}}^{\text{nb}}_{k|k}[ \Delta \mathbf{p}^{\text{b}}_k \times] & \mathbf{0}_{3 \times N_{\text{par}}} \\
        \mathbf{0}_{3 \times 3} & \mathbf{I}_3 & \mathbf{0}_{3 \times N_{\text{par}}} \\
        \mathbf{0}_{N_{\text{par}} \times 3} & \mathbf{0}_{N_{\text{par}} \times 3} & \mathbf{I}_{N_{\text{par}}}
    \end{bmatrix}, \\
    \mathbf{G}_k &= \begin{bmatrix}
        \hat{\mathbf{R}}^{\text{nb}}_{k|k} & \mathbf{0}_{3 \times 3} \\
        \mathbf{0}_{3 \times 3} & \mathbf{I}_3 \\
        \mathbf{0}_{N_{\text{par}} \times 3} & \mathbf{0}_{N_{\text{par}} \times 3}
    \end{bmatrix}, \\
    \mathbf{Q} &= \text{blockdiag}\left(\bm{\Sigma}_{\text{pos}}, \bm{\Sigma}_{\text{rot}}\right),
\end{aligned}
\end{equation}
with $N_{\text{par}} = N_{\text{weights}} + 6N_{\text{mag}}$, where $6N_{\text{mag}}$ is omitted for the algorithm with pre-calibrated magnetometers. Note that we only use the error state for the magnetic field measurement update, not the dynamic update. Instead, we update our state from~\eqref{eq:stateFull} in the dynamic update directly. This effectively means setting the predictive information vector to zero at the start of every magnetic field measurement update.

\subsection{Magnetic field measurement update}
\label{sec:measurementUpdate}
We adapt the measurement model from~\eqref{eq:magMeas} to update the error state from~\eqref{eq:deltaState}, leading to a residual $\mathbf{z}_{k+1}$ as 
\begin{equation}
\begin{aligned}
    \mathbf{z}_{k+1} &= \hat{\mathbf{D}}^{\text{b}^{-1}}_{k+1|k} \mathbf{y}^{\text{b}}_{k+1} -  
     \hat{\mathbf{R}}^{\text{bn}}_{k+1|k}  \nabla\hat{\bm{\Phi}}_{k+1|k} \hat{\mathbf{m}}_{k+1|k} - \hat{\mathbf{b}}^{\text{b}}_{k+1|k}, \\
     \mathbf{z}_{k+1} &\sim \mathcal{N}\left(0, \bm{\Sigma}_{\mathbf{z}_{k+1}}\right),
 \end{aligned}
    \label{eq:measResidual}
\end{equation}
where $\hat{\mathbf{D}}^{\text{b}^{-1}}_{k+1|k} \mathbf{y}^{\text{b}}_{k+1}$ and ${\bm{\Sigma}_{\mathbf{z}_{k+1}} = \hat{\mathbf{D}}^{\text{b}^{-1}}_{k+1|k}  (\hat{\mathbf{D}}^{\text{b}^{-1}}_{k+1|k})^\Transp}$ are the corrected measurement and its noise covariance. We use ${\nabla\hat{\bm{\Phi}}_{k+1|k} = \nabla\bm{\Phi} (\hat{\mathbf{P}}^{\text{n}}_{k+1|k} + \hat{\mathbf{R}}^{\text{nb}}_{k+1|k} \mathbf{S}^{\text{b}} )}$, for notational brevity. The motivation for the corrected measurement model is to improve the linearisation properties compared to using~\eqref{eq:magMeas} directly. Empirically, this yields fewer iterations until convergence. We linearise the measurement equation around the propagated error state $\delta \mathbf{x}_{k+1|k}=0$. The resulting Jacobian is
\begin{equation}
\begin{aligned}
    \mathbf{H}_{i, k+1} &=
    \left[
        \mathbf{H}_{\delta \mathbf{p}_{i,k+1}^{\text{n}}}  \; 
        \mathbf{H}_{\bm{\eta}_{i,k+1}^{\text{b}}}  \; 
        \mathbf{H}_{\delta \mathbf{m}_{i,k+1}}  \; 
        \mathbf{H}_{\delta \mathbf{d}_{i,k+1}^{\text{b}}}  \; 
        \mathbf{H}_{\delta \mathbf{b}_{i,k+1}^{\text{b}}}
    \right], \\
    \mathbf{H}_{k+1} &=
    \left[
        \mathbf{H}_{1, k+1}^\Transp \; \hdots \; \mathbf{H}_{N_{\text{mag}}, k+1}^\Transp
    \right]^\Transp,
    \end{aligned}
    \label{eq:jacobianAll}
\end{equation}
where the notation $\mathbf{H}_{i,k+1}$ denotes row $i$ of the Jacobian corresponding to magnetometer index~$i$. We derive the Jacobians per magnetometer as
\begin{subequations}
\begin{align}
    \mathbf{H}_{\delta \mathbf{p}_{i,k+1}^{\text{n}}} &= -\hat{\mathbf{R}}^{\text{bn}}_{k+1|k} \nabla_{\delta \mathbf{p}} \nabla \hat{\bm{\Phi}}_{i,k+1|k}  \hat{\mathbf{m}}_{k+1|k}, \label{eq:jacobianPosition} \\
    \mathbf{H}_{\bm{\eta}_{i,k+1}^{\text{b}}} &= -\hat{\mathbf{R}}^{\text{bn}}_{k+1|k}   \nabla_{ \bm{\eta}} \nabla \hat{\bm{\Phi}}_{i,k+1|k}\hat{\mathbf{m}}_{k+1|k} \label{eq:jacobianRotation} \\
    &\quad+ \big[ \hat{\mathbf{R}}^{\text{bn}}_{k+1|k} \nabla \hat{\bm{\Phi}}_{i,k+1|k} \hat{\mathbf{m}}_{k+1|k} \times \big], \nonumber  \\
    \mathbf{H}_{\delta \mathbf{m}_{k+1}} &= -\hat{\mathbf{R}}^{\text{bn}}_{k+1|k}  \nabla \hat{\bm{\Phi}}_{i,k+1|k}, \label{eq:jacobianWeights} \\
    \mathbf{H}_{\delta \mathbf{d}_{i,k+1}^{\text{b}}} &= \text{diags}(\mathbf{y}^{\text{b}}_{i,k+1}), \label{eq:jacobianScale} \\
    \mathbf{H}_{\delta \mathbf{b}_{i,k+1}^{\text{b}}} &= -\mathbf{I}_3 \label{eq:jacobianBias},
\end{align}
\end{subequations}
Where $\nabla_{\delta \mathbf{p}} \nabla \hat{\bm{\Phi}}_{i,k+1|k}$ is defined in \citet{viset2022extended} and $\nabla_{\delta \bm{\eta}} \nabla \hat{\bm{\Phi}}_{i,k+1|k}$ is derived using ${\nabla_{\bm{\eta}} \bm{\phi}_j \triangleq \nabla_{\delta \mathbf{p}} \bm{\phi}_j \left[ \mathbf{R}^{\text{nb}}_k\mathbf{s}^{\text{b}}_i \times \right]}$ for basis function $j$. The operator $[\cdot \times]$ is the skew-symmetric cross-product matrix~\citep{kok2017using}. Furthermore, for brevity we use the shorthand notation ${\nabla\hat{\bm{\Phi}}_{i,k+1|k} = \nabla\bm{\Phi} (\hat{\mathbf{p}}^{\text{n}}_{k+1|k} + \hat{\mathbf{R}}^{\text{nb}}_{k+1|k} \mathbf{s}^{\text{b}}_i )}$. We now update the information matrix and information vector as
\begin{subequations}
\label{eq:infoUpdate}
\begin{align}
    \bm{\mathcal{I}}_{k+1|k+1} &= \bm{\mathcal{I}}_{k+1|k} + \mathbf{H}_{k+1} \bm{\Sigma}_{\mathbf{z}_{k+1}}^{-1} \mathbf{H}_{k+1}^\Transp,     \label{eq:infoMatrixUpdate} \\
    \bm{\iota}_{k+1|k+1} &= \mathbf{H}^\Transp_{k+1} \bm{\Sigma}_{\mathbf{z}_{k+1}}^{-1} \mathbf{z}_{k+1}.\label{eq:infoVectorUpdate}
\end{align}
\end{subequations}
where we used the fact that the predictive information vector is zero. The posterior error state is then found as
\begin{equation}
    \delta \hat{\mathbf{x}}_{k+1|k+1} =  \bm{\mathcal{I}}_{k+1|k+1}^{-1} \bm{\iota}_{k+1|k+1}, 
    \label{eq:posteriorFromInfo}
\end{equation}
which we use to update our prior state estimate to the posterior as 
\begin{equation}
    \hat{\mathbf{x}}_{k+1|k+1} = \hat{\mathbf{x}}_{k+1|k} \oplus \delta \hat{\mathbf{x}}_{k+1|k+1},
    \label{eq:measUpdate}
\end{equation}
where the following operations define $\oplus$ 
\begin{subequations}
\label{eq:measuremenUpdates}
\begin{align}   
    \hat{\mathbf{p}}^{\text{n}}_{k+1|k+1} &= \hat{\mathbf{p}}^{\text{n}}_{k+1|k} + \delta \hat{\mathbf{p}}^{\text{n}}_{k+1|k+1},  \label{eq:measuremenUpdatePosition} \\
    \hat{\mathbf{R}}^{\text{bn}}_{k+1|k+1} &= \text{exp}_{\mathbf{R}}\left( \hat{\bm{\eta}}^{\text{b}}_{k+1|k+1}\right)\hat{\mathbf{R}}^{\text{bn}}_{k+1|k}, \label{eq:measuremenUpdateRotation} \\
    \hat{\mathbf{m}}_{k+1|k+1} &= \hat{\mathbf{m}}_{k+1|k} + \delta \hat{\mathbf{m}}_{k+1|k+1}, \label{eq:measuremenUpdateWeights}  \\
    \hat{\mathbf{d}}^{\text{b}}_{k+1|k+1} &= \hat{\mathbf{d}}^{\text{b}}_{k+1|k} + \delta \hat{\mathbf{d}}^{\text{b}}_{k+1|k+1}, \label{eq:measuremenUpdateScale}  \\
    \hat{\mathbf{b}}^{\text{b}}_{k+1|k+1} &= \hat{\mathbf{b}}^{\text{b}}_{k+1|k} + \delta \hat{\mathbf{b}}^{\text{b}}_{k+1|k+1}.\label{eq:measuremenUpdateBias}
\end{align}
\end{subequations}

\subsection{Vertical position measurement update}
We found that vertical drift negatively affected the ability to perform loop closures, severely degrading the algorithm's performance. To compensate for this, we introduce a vertical position pseudo-measurement update, based on the ground truth data, between the dynamic update and the magnetic field measurement update. The residual $\mathbf{z}_{\text{ver},k+1}$ for this update is found as
\begin{equation}
\begin{aligned}
    \mathbf{z}_{\text{ver},k+1} &= \mathbf{y}_{\text{ver},k+1} - \mathbf{H}_{\text{ver}}\delta\hat{\mathbf{x}}_{k+1|k}, \\
    \mathbf{z}_{\text{ver},k+1} &\sim \mathcal{N}\left( 0, \bm{\Sigma}_{\text{ver}}\right),
\end{aligned}
\label{eq:residualVertical}
\end{equation}
where $\bm{\Sigma}_{\text{ver}}$ is the pseudo-measurement covariance, $\mathbf{y}_{\text{ver},k+1}$ is the (assumed known) ground truth vertical position and $\mathbf{H}_{\text{ver}} \delta \hat{\mathbf{x}}_{k+1|k} \triangleq \hat{p}^{\text{n}}_{3,k+1|k}$ is the predicted vertical position after the dynamic update. This is then used to update the state using \eqref{eq:infoUpdate}, \eqref{eq:posteriorFromInfo} and \eqref{eq:measuremenUpdates} similarly to the magnetic field measurement update.

\subsection{SL(C)AMma algorithm}

As we do SLAM, we assume, without loss of generality \citep{gustafsson2010statistical}, that the initial position and orientation are known, which leads to the initial information matrix of the error state as
\begin{equation}
    \bm{\mathcal{I}}_{0|0} = \text{blockdiag}\left(\mathbf{I}_{3} 10^{12}, \mathbf{I}_{3} 10^{12}, \bm{\Lambda}^{-1}_{\text{w}},\bm{\Lambda}^{-1}_{\text{d}}, \bm{\Lambda}^{-1}_{\text{b}}\right),
    \label{eq:initCovariance}
\end{equation}
where $\bm{\Lambda}_{\text{w}}$ is defined in \eqref{eq:priorWeights}. Additionally, $\bm{\Lambda}_{\text{d}}$ and $\bm{\Lambda}_{\text{b}}$ are the prior covariances of $\mathbf{d}^{\text{b}}$ and $\mathbf{b}^{\text{b}}$ and are only added in the SLCAMma framework. We summarise the algorithm for both SLAMma and SLCAMma in Algorithm \ref{alg:SCLAM}. We define convergence for the iterated magnetic field measurement update as ${\|\delta \hat{\mathbf{p}}^{\text{n}}_{k+1|k+1}\| < 0.1}$ mm and ${\|\hat{\bm{\eta}}^{\text{b}}_{k+1|k+1}\| < 0.1^\circ}$ and allow for a maximum of ${\tau_{\text{max}} = 5}$ iterations. We focus on convergence in terms of the change in position and orientation, as localisation is the main interest.  

\begin{algorithm}
\caption{Simultaneous Localisation (Calibration) and Mapping using a magnetometer array}
\label{alg:SCLAM}
\begin{algorithmic}[1]
    \State \textbf{Input:} $ \{\mathbf{y}^\text{b}_{k}, \Delta \mathbf{p}^{\text{b}}_k, \Delta \mathbf{R}_k \}_{k=1}^{N_{\text{K}}}$
    \State \textbf{Output:} $\{\hat{\mathbf{x}}_{k|k}, \bm{\mathcal{I}}_{k|k}\}_{k=1}^{N_{\text{K}}}$
    \State \textbf{Initialise:} $\hat{\mathbf{p}}^{\text{n}}_{0|0} = \mathbf{0}_3$, $\hat{\mathbf{R}}^{\text{bn}}_{0|0} = \mathbf{I}_3$, $\hat{\mathbf{m}}_{0|0} = \mathbf{0}_{N_{\text{weights}}}$
    \State \quad  and $\bm{\mathcal{I}}_{0|0}$ using \eqref{eq:initCovariance}.
    \State \quad \textit{If calibration included:} $\hat{\mathbf{d}}^{\text{b}}_{0|0} = \mathbf{1}_{3N_{\text{mag}}}$, $\hat{\mathbf{b}}^{\text{b}}_{0|0} = \mathbf{0}_{3N_{\text{mag}}}$.
    \For{$k = 0$ to $N_{\text{K}}-1$}
        \State \textbf{Dynamic update:} 
        \State \quad Compute $\hat{\mathbf{x}}_{k+1|k}$ using \eqref{eq:dynamicUpdates}.
        \State \quad Compute $\bm{\mathcal{I}}_{k+1|k}$ using \eqref{eq:dynamicInformationUpdate}.
        \State \textbf{Vertical position pseudo-measurement update:} 
        \State \quad Update $\hat{\mathbf{x}}_{k+1|k}$ and $\bm{\mathcal{I}}_{k+1|k}$ using \eqref{eq:infoUpdate}-\eqref{eq:residualVertical}.
        
        \State \textbf{Magnetometer array measurement update:}
        \State \quad Initialise iteration: $\hat{\mathbf{x}}^{(\tau)}_{k+1} \gets \hat{\mathbf{x}}_{k+1|k}$, $\tau \gets 0$.
        \Repeat
        \State Compute $\mathbf{H}_{k+1}^{(\tau)}$ at $\hat{\mathbf{x}}^{(\tau)}_{k+1}$ using \eqref{eq:jacobianAll}.
        \State Compute $ \delta \hat{\mathbf{x}}^{(\tau)}_{k+1} $ using \eqref{eq:posteriorFromInfo} with 
        \State \quad $\bm{\mathcal{I}}^{(\tau)}_{k+1|k+1}$ and $\bm{\iota}^{(\tau)}_{k+1|k+1}$ by using \eqref{eq:infoUpdate} 
        \State \quad with $\mathbf{H}_{k+1}^{(\tau)}$, $\bm{\Sigma}_{\mathbf{z}_{k+1}}^{-1}$ and $\bm{\mathcal{I}}_{k+1|k}$. 
        \State Update $\hat{\mathbf{x}}^{(\tau+1)}_{k+1} \gets \hat{\mathbf{x}}^{(\tau)}_{k+1} \oplus \delta \hat{\mathbf{x}}^{(\tau)}_{k+1}$ using 
            \eqref{eq:measuremenUpdates}. 
        \State $\tau \gets \tau + 1$.
        \Until{convergence or $\tau = \tau_{\text{max}}$}.
        \State \textbf{Store variables:} 
        \State \quad $\hat{\mathbf{x}}_{k+1|k+1} \gets \hat{\mathbf{x}}^{(\tau)}_{k+1}$, $\bm{\mathcal{I}}_{k+1|k+1} \gets \bm{\mathcal{I}}^{(\tau)}_{k+1|k+1}$.
    \EndFor
\end{algorithmic}
\end{algorithm}
\section{Simulation results on the convergence of the calibration parameter estimates}
\label{sec:Simulations}
To analyse the convergence of the estimated calibration parameters to the true calibration parameters, we simulate a magnetometer array moving in different ways through a disturbed magnetic field. We assume a 30-magnetometer array of the same configuration as in Figure~\ref{fig:rigidBody}. We sample scaling factors and biases as ${\{\{d_{n_d,i}\}_{i=1}^{N_{\text{mag}}}\}_{n_d=1}^{3} \sim \mathcal{U}(0.9,1.1)}$ and ${\{\{b_{n_d,i}\}_{i=1}^{N_{\text{mag}}}\}_{n_d=1}^{3} \sim \mathcal{U}(-1.5,1.5)}$ respectively, for every dimension $n_d$. Here, $\mathcal{U}(\cdot,\cdot)$ denotes the uniform distribution. We simulate the constant Earth's magnetic field as a vector $\begin{bmatrix} 19.2 &
0.8 &
45.5
\end{bmatrix}^\Transp \mu \text{T}$, approximately equal to its value in Delft, the Netherlands. On top of this constant field, we simulate magnetic field disturbances by sampling from a GP with the curl-free assumption from~\cite{wahlstrom2013modeling}, and hyper-parameters $\sigma_{\text{cf}} = 1$, $\ell = 1$ and $\sigma_{\text{y}} = 0.1$. We use the same hyper-parameters in our reduced rank GP, with additionally $\sigma_{\text{lin}} = 15$.  The chosen settings for the simulation are reflective of those used in the experiments in Section \ref{sec:experiments}.

We impose strong priors of $\bm{\Lambda}_{\text{d}} = 0.001^2 \mathbf{I}_{3N_{\text{mag}}}$ on the calibration scaling parameters to prevent filter divergence in the early time-steps. The prior on the bias is set as $\bm{\Lambda}_{\text{b}} = 0.1^2 \mathbf{I}_{3N_{\text{mag}}}$. We assume a dynamic model as described in \eqref{eq:dynamicModel} with covariances ${\bm{\Sigma}_{\text{pos}}^{1/2} = 0.01 \mathbf{I}_3 \mathrm{m}s^{-1}}$ and ${\bm{\Sigma}_{\text{rot}}^{1/2} =  0.1 \mathbf{I}_3 \mathrm{deg} \mathrm{s}^{-1}}$, at a frequency of $10\mathrm{Hz}$.

\begin{figure*}[t!]
    \centering
    \includegraphics[width=\linewidth]{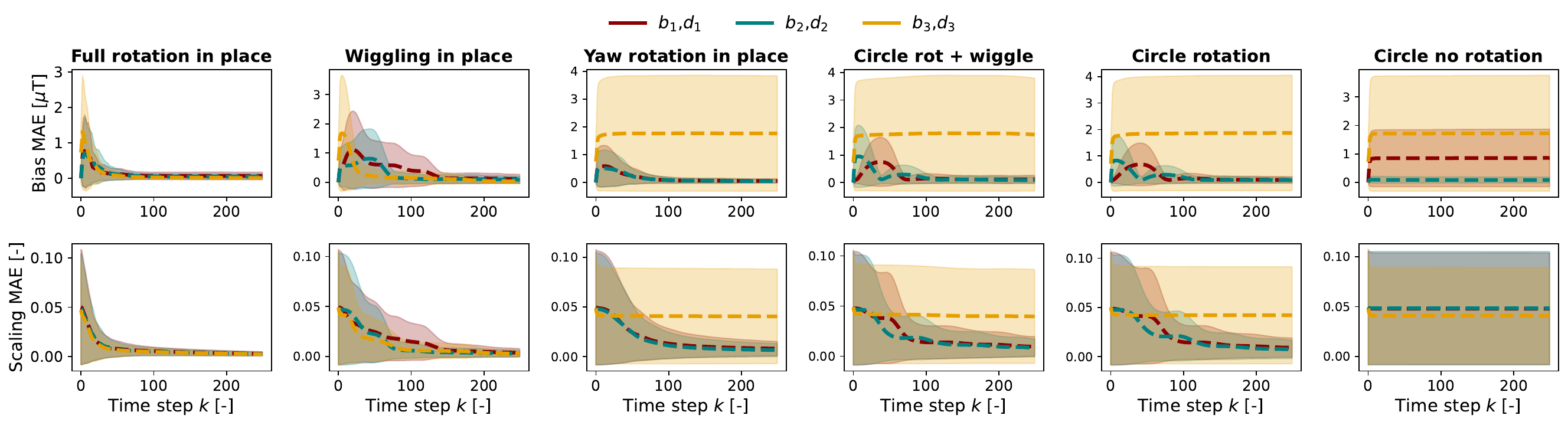}
    \caption{MAE and 1 STD of the estimated bias and scaling parameters across 30 MC simulations for six types of motion.}
    \label{fig:calibrationIdentifiability}
\end{figure*}

As a metric for analysis, we use the mean absolute error (MAE) and 1 standard deviation (STD) of the bias $\mathbf{b}^{\text{b}}_{i,k}$ and scaling error $\mathbf{d}^{\text{b}}_{i,k}$ for different types movement over all magnetometers and Monte Carlo (MC) simulations, which are calculated as
\begin{equation}
\begin{aligned}
    \text{MAE:} \quad \mu_{\bm{\theta},k}
    &= \frac{1}{N_{\text{mc}} N_{\text{mag}}} \sum_{i=1}^{N_{\text{mag}}} \sum_{j=1}^{N_{\text{mc}}} \left| \hat{\bm{\theta}}_{i,j,k|k}^{\text{b}} - \bm{\theta}_{i,j}^{\text{b}} \right|, \\
    \text{STD:} \quad \sigma_{\bm{\theta},k}
    &= \sqrt{\frac{1}{N_{\text{mc}} N_{\text{mag}}} \sum_{i=1}^{N_{\text{mag}}} \sum_{j=1}^{N_{\text{mc}}} \left\| \hat{\bm{\theta}}_{i,j,k|k}^{\text{b}} - \bm{\theta}_{i,j}^{\text{b}} \right\|_2^2}.
\end{aligned}
\label{eq:meanStdCalibrationIdentifiability}
\end{equation}
Here, $|\cdot|$ denotes the absolute value, $||\cdot||_2$ denotes the Euclidean norm, and we use the placeholder variable ${\bm{\theta} \in \{\mathbf{d}, \mathbf{b} \}}$ to denote the scaling vector or the bias. Finally, $N_{\text{mc}}$ is the number of MC simulations. 

We analyse the convergence of the calibration parameters across different simulated motion types: The magnetometer array rotates randomly in place to explore as much of the orientation space as possible in \texttt{Full rotation in place}. The magnetometer rotates in place with $5^\circ$ around all angles in \texttt{Wiggling in place}. Next, the magnetometer array rotates back and forth around the yaw axis in \texttt{Yaw rotation in place}. Further, the magnetometer array translates along a planar circle of approximately 0.1m radius, without rotating, in \texttt{Circle no rotation}. Next, the magnetometer array moves along a planar circle, but aligns the heading angle with the direction of movement, in \texttt{Circle yaw rotation}. Finally, the magnetometer array moves the same as in the previous motion, while also wiggling $5^\circ$ around the pitch and roll angle, in \texttt{Circle wiggle}. 

\begin{figure}[h!]
    \centering
    \includegraphics[width=\linewidth]{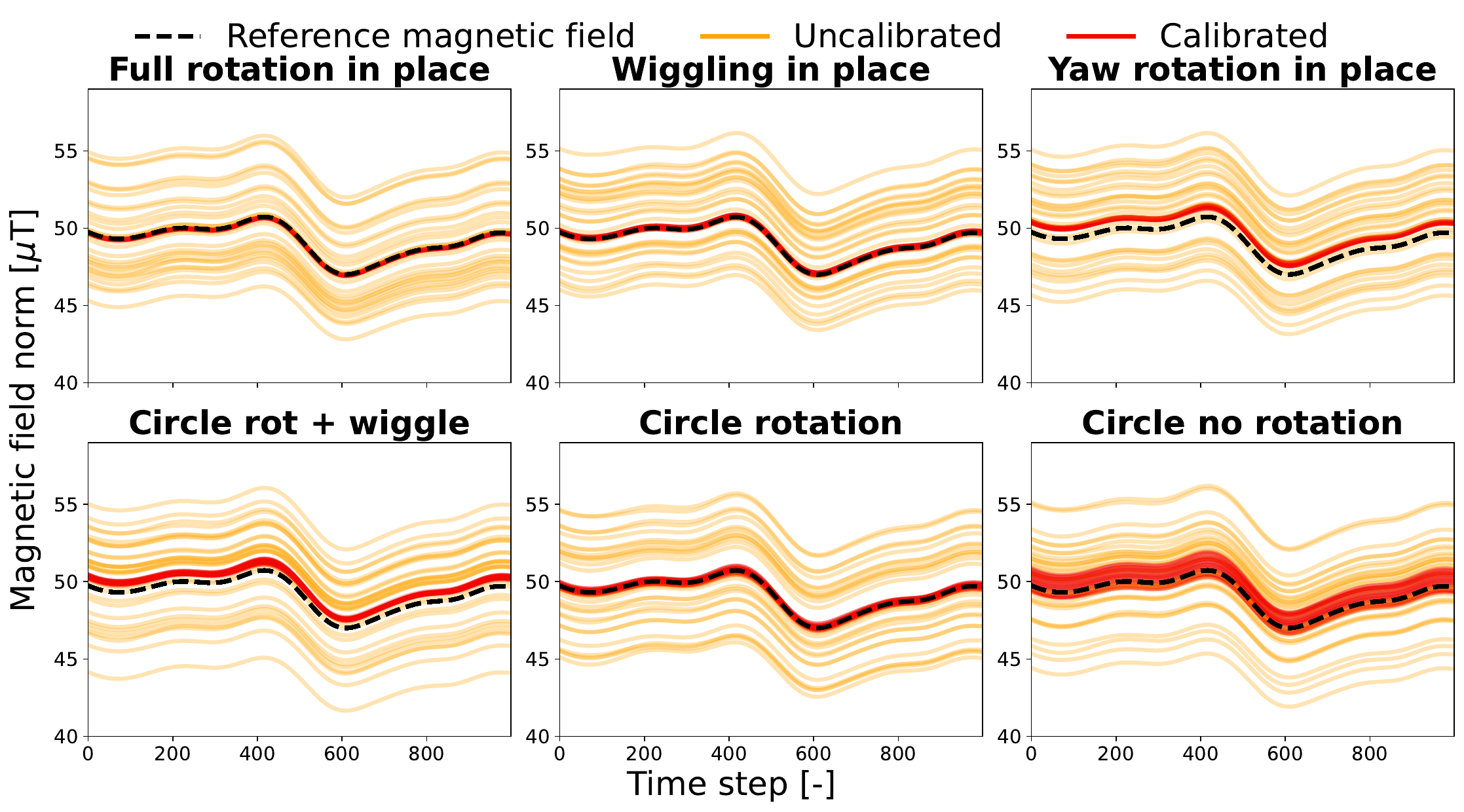}
    \caption{Demonstration of the inter-sensor measurement consistency for all six simulations. Calibrated norms are compared to uncalibrated norms and a reference magnetic field.}
    \label{fig:calibrationSimilarityExample}
\end{figure}

The first two columns of Figure~\ref{fig:calibrationIdentifiability} show that larger orientation excitation improves calibration parameter estimation, where the estimation errors approach zero even with minimal rotation ($5^\circ$ for all axes). In contrast, columns 3-5 demonstrate that for planar motion, neither the vertical bias $b_3$ nor the scaling factor $d_3$ converge to the true values. The MAE of $b_3$ remains approximately at 0.9~$\mu$T, while the MAE of $d_3$ remains constant around 0.02. This is in line with~\cite{vallivaara2024saying}, where no using only planar motion, they obtain no information on the vertical bias component. Finally, the last column shows that with translation only, the bias estimates fail to converge to the true values. The MAE grows to approximately $[0.2, 0.4, 0.9]~\mu\mathrm{T}$ as the filter optimises for inter-sensor measurement consistency across the magnetometers. We further analyse this effect in Figure~\ref{fig:calibrationSimilarityExample}.

Inter-sensor measurement consistency across the magnetometers is critical for our SLAM algorithm. Specifically, magnetometers in the same position and orientation should measure the same values. We validate this consistency across the six different motion types by comparing the measurements norms of the uncalibrated magnetometers against the calibrated magnetometers using a new planar motion dataset. We generate a new synthetic magnetic field and assume that all magnetometers experience the same reference field for every time-step $k$. We then apply the bias and scaling factors from the respective simulation to get uncalibrated measurements and apply the final online calibration parameters from Figure~\ref{fig:calibrationIdentifiability} to get the calibrated measurements. The results, shown in Figure~\ref{fig:calibrationSimilarityExample}, demonstrate that inter-sensor measurements converge to consistent values within any given movement type, in this case, planar motion. While limited orientation excitation (e.g., \texttt{Circle no rotation}) produces consistent results with a range of~$1.5 \mu \mathrm{T}$, higher orientation excitation (e.g., \texttt{Full rotation in place}) results in a higher consistency, showing a lower range of~$0.1 \mu \mathrm{T}$.

\section{Experimental results}
\label{sec:experiments}
We also study the performance of our algorithms on experimental datasets collected using a 30-magnetometer array\footnote{Available online at \href{https://hendeby.gitlab-pages.liu.se/research/magdata/}{https://hendeby.gitlab-pages.liu.se/research/magdata/}}. The magnetometer array has 30 PNI RM3100 magnetometers sampling at 100~Hz, arranged in a planar configuration as shown in Figure~\ref{fig:rigidBody}. Ground truth position and orientations are available, provided by an OptiTrack motion capture system. We use constant hyper-parameters $\sigma_{\text{y}} = 0.1$ and $\sigma_{\text{lin}} = 15$ and the additional hyper-parameters $\{\ell, \sigma_{\text{se}}\}$ per scenario are shown in Table~\ref{tab:experimentSettings}. The length-scale $\ell$ and signal variance $\sigma_{\text{se}}$ for the first two datasets are the same as reported in~\cite{edridge2025position}, and are optimised for the rest of the scenarios, according to~\cite{solin2018modeling}. 

The changes in position and orientation used in~\eqref{eq:dynamicModel} are calculated from the ground truth data. To simulate drift, we add a constant term to both the change in position and orientation as ${\Delta \mathbf{p}^{\text{n}}_{k} = \Delta \bar{\mathbf{p}}^{\text{n}}_{k} + \mathbf{o}_{\text{pos}}}$ and ${\Delta \mathbf{R}_{k} = \text{exp}_{\mathbf{R}}(\mathbf{o}_{\text{rot}})\Delta \bar{\mathbf{R}}_{k}}$, where ${\bar{\mathbf{p}}^{\text{n}}_{k}}$ and $\Delta \bar{\mathbf{R}}_{k}$ denote the true change in position and orientation, respectively. The drift parameter values are listed in Table~\ref{tab:experimentSettings}, along with the number of basis functions and noise covariances. The prior is~\eqref{eq:initCovariance} with ${\bm{\Lambda}_{{\text{d}}} = 0.001 \mathbf{I}_{3N_{\text{mag}}} }$ and ${\bm{\Lambda}_{{\text{b}}} =  0.1 \mathbf{I}_{3N_{\text{mag}}}}$, the same values as used in the simulations from Section~\ref{sec:Simulations}. Similar to the simulations, it was found that the strong prior on the scaling parameters helps prevent filter divergence in the early time-steps. We run both versions Algorithm~\ref{alg:SCLAM} and compare this against single magnetometer SLAM from~\cite{viset2022extended}, using the same settings and with a vertical position update, but restricting the measurements to only one magnetometer on the array. We also compare the results to dead reckoning, which uses the dynamic model from~\eqref{eq:dynamicModel}, without corrections from the measurement updates.

\begin{table*}[t]
\centering
\caption{Settings used in the experiments.}
\resizebox{\textwidth}{!}{%
\begin{tabular}{lcccccccc}
\toprule
\textbf{Scenario} 
    & Time $[\mathrm{s}]$
    & Length $[\mathrm{m}]$
    & $\ell$ $[\mathrm{m}]$
    & $\sigma_{\text{se}}/\ell$ [$\mu \mathrm{T} m^{-1}$]
    & $\bm{\Sigma}_{\mathrm{pos}}^{1/2}$ [$\mathrm{mm \,s}^{-1}$]
    & $\bm{\Sigma}_{\mathrm{rot}}^{1/2}$ [$\mathrm{deg \,s}^{-1}$]
    & $\mathbf{o}_{\text{pos}}$ [$\mathrm{mm\,s}^{-1}$]
    & $\mathbf{o}_{\text{rot}}$ [$\mathrm{deg\,s}^{-1}$] 
\\
\midrule
\textbf{Tiny no rot} & 40.5 & 6.15 & 0.15 & 5 & $10\mathbf{I}_3$ & $\mathrm{diag}(0.1,0.1,1)$ & $[50\ -50\ 0]^\Transp$ & $[0\ 0\ -1]^\Transp$ \\
\textbf{Tiny yaw rot} & 96.6 & 6.23 & 0.15 & 5 & $10\mathbf{I}_3$ & $\mathrm{diag}(0.1,0.1,1)$ & $[50\ -50\ 0]^\Transp$ & $[0\ 0\ -1]^\Transp$ \\
\textbf{Snake wide 1} & 82.3 & 52 & 0.5 & 1.25 & $10\mathbf{I}_3$ & $\mathrm{diag}(0.1,0.1,1)$ & $[-50\ 50\ 0]^\Transp$ & $[0\ 0\ 1]^\Transp$ \\

\textbf{Snake wide 2} & 70.7 & 50.5 & 0.85 & 1.78 & $10\mathbf{I}_3$ & $\mathrm{diag}(0.1,0.1,1)$ & $[-25\ 25\ 0]^\Transp$ & $[0\ 0\ 0.5]^\Transp$ \\

\textbf{Squares short} & 67.1 & 45.4 & 0.912 & 1.88 & $10\mathbf{I}_3$ & $\mathrm{diag}(0.1,0.1,1)$ & $[-50\ 50\ 0]^\Transp$ & $[0\ 0\ 1]^\Transp$ \\
\textbf{Squares long} & 98.2 & 63.8 & 0.975 & 2.19 & $10\mathbf{I}_3$ & $\mathrm{diag}(0.1,0.1,1)$ & $[25\ -25\ 0]^\Transp$ & $[0\ 0\ -0.5]^\Transp$ \\
\textbf{Snake long} & 164 & 133 & 0.712 & 2.4 & $10\mathbf{I}_3$ & $\mathrm{diag}(0.1,0.1,1)$ & $[-50\ 50\ 0]^\Transp$ & $[0\ 0\ 1]^\Transp$ \\
\textbf{Snake thin 1} & 88.5 & 62.4 & 0.9 & 0.9 & $10\mathbf{I}_3$ & $\mathrm{diag}(0.1,0.1,1)$ & $[25\ -25\ 0]^\Transp$ & $[0\ 0\ -0.5]^\Transp$ \\
\textbf{Snake thin 2} & 82.1 & 63.4 & 0.9 & 0.9 & $10\mathbf{I}_3$ & $\mathrm{diag}(0.1,0.1,1)$ & $[-25\ 25\ 0]^\Transp$ & $[0\ 0\ 0.5]^\Transp$ \\
\textbf{Infinity symbol} & 57.2 & 68.2 & 1.07 & 3.36 & $10\mathbf{I}_3$ & $\mathrm{diag}(0.1,0.1,1)$ & $[50\ -50\ 0]^\Transp$ & $[0\ 0\ -1]^\Transp$ \\
\bottomrule
\end{tabular}%
}
\label{tab:experimentSettings}
\end{table*}

\begin{figure}[h!]
    \centering
    \includegraphics[width=\linewidth]{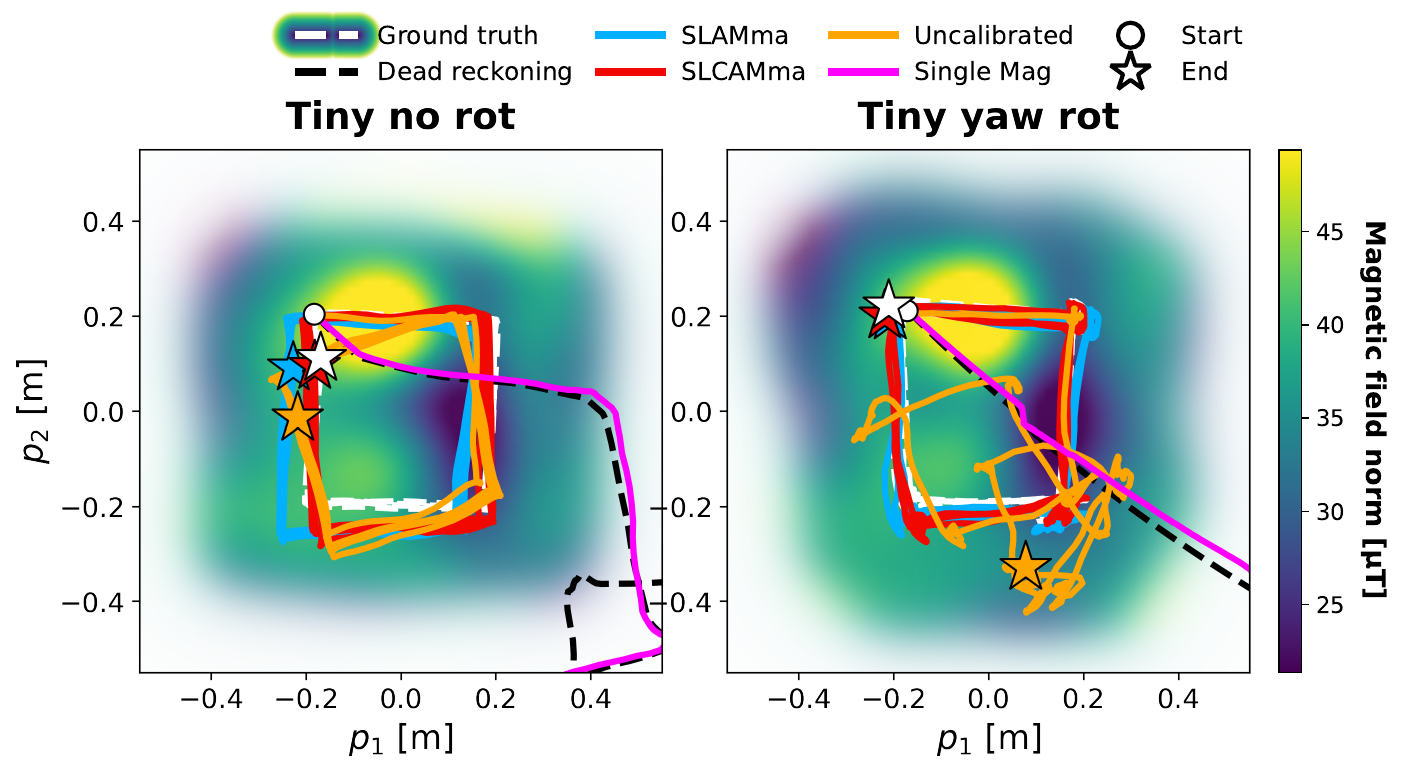}
    \includegraphics[width=\linewidth]{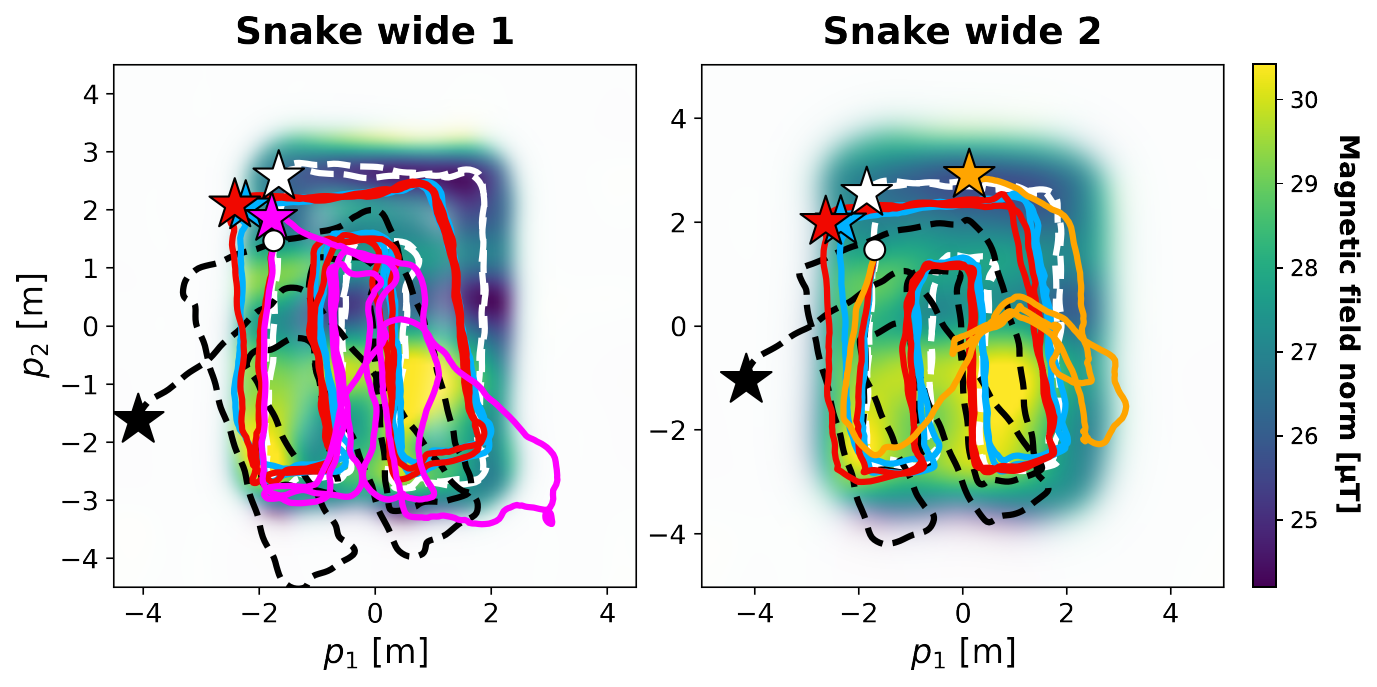}
    \caption{Estimated trajectories for the (top) \texttt{tiny square} and (bottom) \texttt{small snakes} experiments, using a 30-magnetometer array. The background shows the predicted magnetic field norm based on the ground truth positions, and the transparency is proportional to the prediction uncertainty.}
    \label{fig:tinySquareAndSmallSnakes}
\end{figure}

Figure~\ref{fig:tinySquareAndSmallSnakes} (top) shows the \texttt{Tiny no rot} (only translation) and \texttt{tiny yaw rot} ($90^\circ$ corner turns) datasets. Both were recorded in a tiny square in close proximity to magnetic field disturbances, causing a high magnetic field norm range of $[20,50]\mu \mathrm{T}$. The estimated trajectories for \texttt{Tiny no rot} demonstrate that the magnetometer array (SLAMma/SLCAMma), reduces positional and rotational drift significantly by approximately $7.0 \, \mathrm{cm \, s^{-1}}$ and $1.0^\circ \, \mathrm{s^{-1}}$, respectively, compared to dead reckoning and $4.2 \, \mathrm{cm \, s^{-1}}$ and $1.6^\circ \, \mathrm{s^{-1}}$ compared to single magnetometer SLAM. While an uncalibrated array produces similar estimates for only translation, it fails to enable a loop-closure after the array has rotated. This is a result of a mismatch between the created magnetic field map and the measurements after rotating, such that when maximising the posterior \eqref{eq:posteriorPDF}, the array is repelled rather than attracted to the area of the loop-closure. These results highlight the strong potential of using using a magnetometer array and demonstrate the importance of (online) magnetometer calibration for SLAM. 
 
The wide snake trajectories at the bottom of Figure~\ref{fig:tinySquareAndSmallSnakes} demonstrate a more realistic scenario, with a significantly larger domain and lower signal-to-noise ratio (SNR), resulting in a longer and more difficult exploration phase before loop closures. Both trajectories follow a snake-like pattern without overlap before they loop back
to the starting location. Both have a magnetic field norm range of approximately $[24,30]\mu \mathrm{T}$. In contrast to the tiny squares, SLAMma with uncalibrated magnetometers diverges after the second corner, as shown on the right, where both SLAMma (with calibrated magnetomters) and SLCAMma follow the ground-truth trajectory. This again demonstrates the importance of magnetometer calibration for navigation purposes. 

\begin{figure}[h!]
    \centering
    \includegraphics[width=\linewidth]{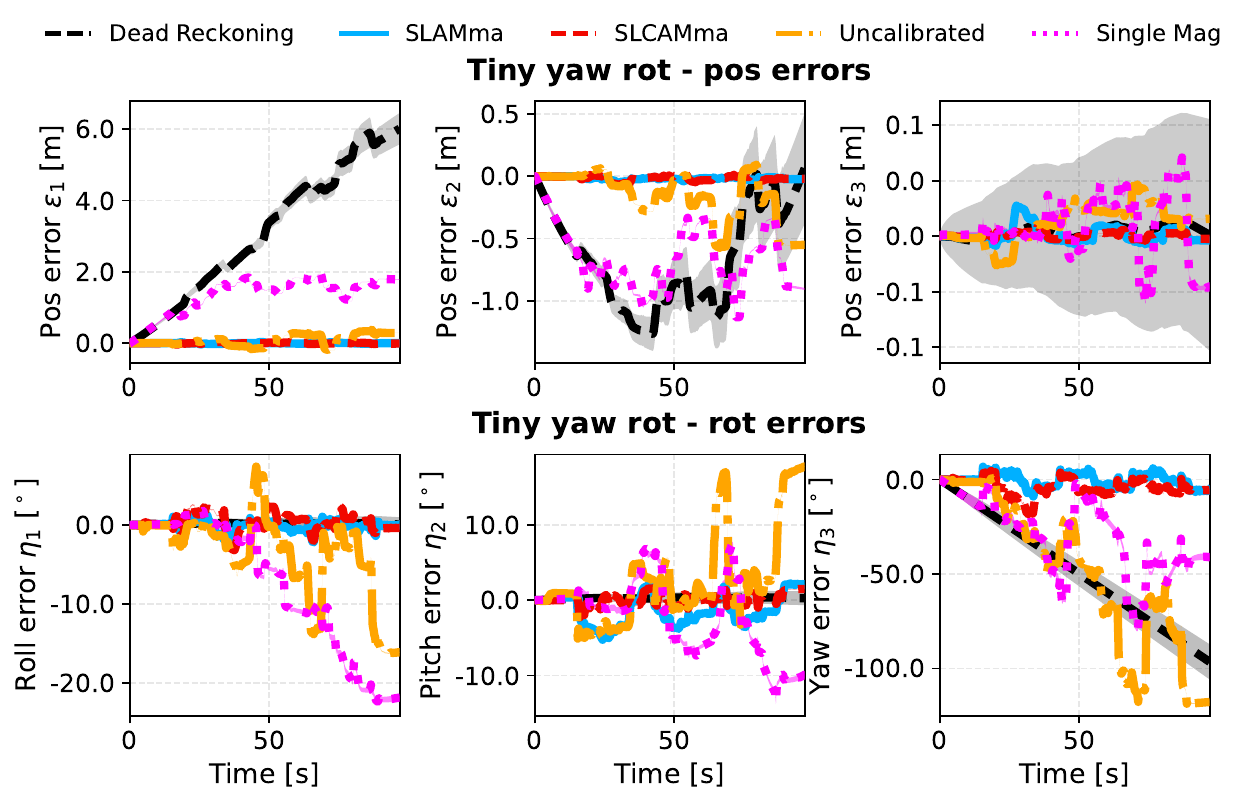}
    \caption{Position and rotation errors for the \texttt{Tiny square yaw rot} experiment, using a 30-magnetometer array. The estimation uncertainties of three standard deviations are too small to be shown, except for dead reckoning.}
    \label{fig:errorTinySquare}
\end{figure}

We also demonstrate in both \texttt{Tiny} scenarios the single magnetometer SLAM from~\cite{viset2022extended} fails to correct for the amount of drift present and largely follows the dead reckoning trajectory in both \texttt{Tiny} scenarios. This is shown in more detail in the error plots in Figure~\ref{fig:errorTinySquare}, where, initially, the single magnetometer SLAM closely follows the dead reckoning position and rotation error, resulting in errors too large to successfully perform a loop-closure. These results clearly highlight the benefit of using multiple magnetometers over a single magnetometer in the presence of magnetic fields with high disturbances and small length-scales, as it benefits from the odometric properties as shown in \citep{skog2018magnetic, skog2021magnetic, edridge2025position, skog2025connection}. 

\begin{figure}[h!]
    \centering
    \includegraphics[width=\linewidth]{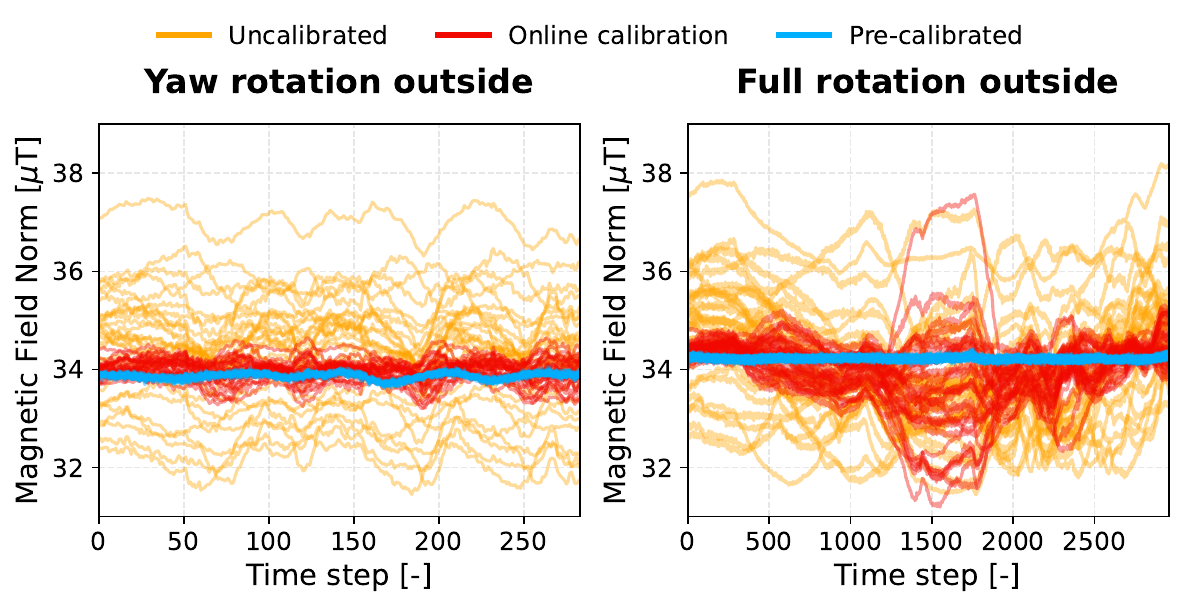}
    \caption{Magnetometer measurements of validation experiments using the final calibration parameters from the \texttt{Small snake high} experiment (Figure~\ref{fig:tinySquareAndSmallSnakes}). Left: yaw rotation outdoors. Right: rotation around all axes, outdoors.}
    \label{fig:calibrationExperimentComparison}
\end{figure}

To verify whether the online calibration is successful and the magnetometers in the SLCAMma algorithm reach inter-sensor measurement consistency, similar to the results in Figure~\ref{fig:calibrationSimilarityExample}, we compare the magnetometer measurements on two new datasets collected outdoors. This is far away from any magnetic field disturbances, resulting in approximately only the Earth's magnetic field. The first dataset considers only yaw rotations, while the second dataset rotates around all axes. Instead of comparing to the ground truth, we compare to Earth's magnetic field of pre-calibrated magnetometers. We take the calibration parameters from SLCAMma at the end of the \texttt{Small snake high} trajectory. We compare three different cases, first is uncalibrated magnetometers, second, is with pre-calibrated magnetometers and third, using the online calibrated magnetometers from the SLCAMma \texttt{Small snake high} experiment, as shown in Figure~\ref{fig:tinySquareAndSmallSnakes}. We show the results in Figure~\ref{fig:calibrationExperimentComparison}. The measurements of online calibration with only yaw rotations congregate around $34 \mu\mathrm{T}$, with approximately $0.5 \mu\mathrm{T}$ more deviation compared to the pre-calibrated magnetometers, showing similar performance results to the simulations from Section~\ref{sec:Simulations}. The results also show no consensus between calibration parameters for non-yaw rotations. This is because, as shown in Section~\ref{sec:Simulations}, the vertical calibration parameters errors don't converge to zero when only yaw rotations are used. 

For the remainder of the experiments, we consider scenarios with longer trajectories and more complicated loop closures, i.e., loop closures after longer time periods and at different angles. The datasets are \texttt{Squares short} and \texttt{Squares long}, where the trajectories follow 4 overlapping squares, which come in from different sides. Furthermore, \texttt{Snake thin 1} and \texttt{Snake thin 2} both follow a snake-like trajectory without overlap, which then loops back to the start. Additionally, \texttt{Snake long} first follows a snake-like pattern without crossing its own path, after which it follows a snake-like pattern at an angle of $90^\circ$ continuously crossing its own path. Lastly, \texttt{Infinity symbol} follows the shape of infinity crossing itself in the middle with a $180^\circ$ angle. All six datasets are collected at approximately hip height (except for \texttt{snake thin 1} at knee height).

\begin{figure*}[t!]
    \centering
    \includegraphics[width=1\linewidth]{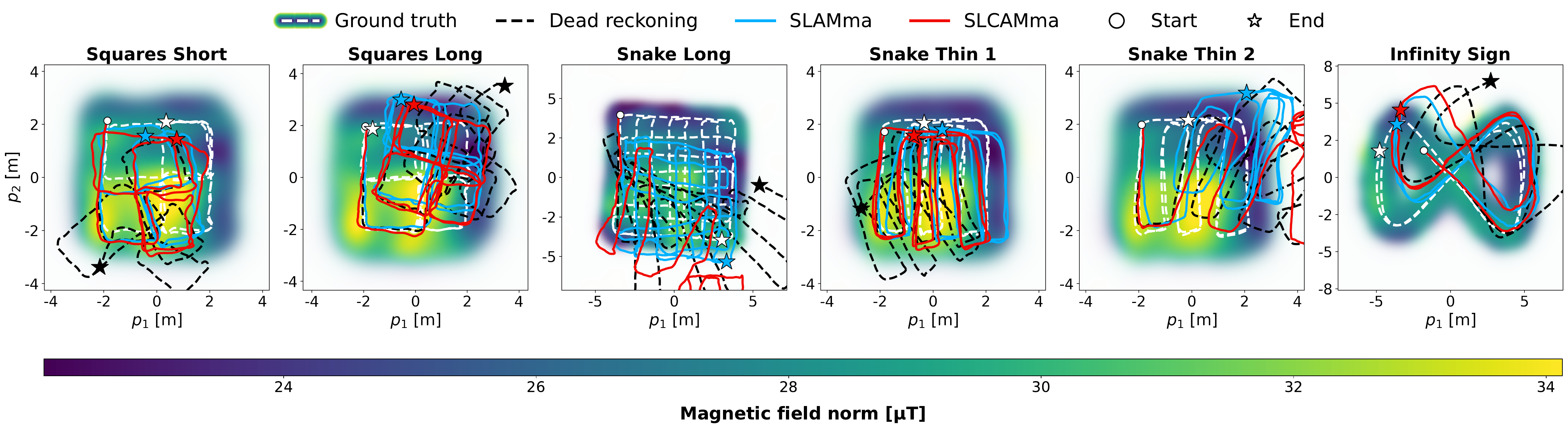}
    \caption{Estimated trajectories for six longer experiments, using a 30-magnetometer array. The background colour shows the predicted magnetic field norm, and the transparency is proportional to the prediction uncertainty.}
    \label{fig:driftMajorComparison}
\end{figure*}

The results are plotted in Figure~\ref{fig:driftMajorComparison}. For clarity of visualisation, we do not show the single magnetometer SLAM and uncalibrated SLAMma as these gave poor performances, except for single magnetometer SLAM in \texttt{Squares long} and \texttt{Snake thin 1}. Overall, the results again highlight the efficacy of using a magnetometer array for SLAM, even in mildly disturbed magnetic field areas. For instance in \texttt{Squares short}, the algorithms $\{\text{SLAMma},\text{SLCAMma}\}$ reduce up to approximately $\{84, 87\}\%$ positional drift and $\{92, 72\}\%$ rotational drift compared to dead reckoning, respectively. However, in the case of longer exploratory phases and significant drift, in \texttt{Snake long} and \texttt{Snake thin 2}, SLCAMma is unable to perform a loop closure, and the filter diverges. This highlights the limitations of online magnetometer calibration, as it can reduce the robustness for longer time periods between loop closures or for lower quality dynamic models.

\section{Conclusion and future work}
\label{sec:conclusion}
This paper demonstrates the effectiveness of magnetometer array SLAM using both pre-calibrated and online magnetometer calibration. Both algorithms produce similar trajectories and significantly reduce drift, particularly in areas with highly disturbed magnetic fields. This shows the superiority of using magnetometer arrays over single magnetometers, which robots can easily carry on their platforms. While due to inter-sensor measurement inconsistency the online calibration filter can diverge during an extensive exploration before visiting a previously seen location, we demonstrate that sufficient orientation excitation reduces this issue. Interesting ideas for future work include extending the proposed algorithms with an accelerometer and gyroscope or selectively applying online magnetometer calibration only in highly disturbed areas.

\begin{dci}
The author(s) declare no potential conflicts of interest with respect to the research, authorship, and/or publication of this article.
\end{dci}

\begin{acks}
This work is funded by the Sensor AI Lab, under the AI Labs program of Delft University of Technology. We thank Gustaf Hendeby from Link\"oping University, and Isaac Skog and Chuan Huang from KTH Royal Institute of Technology for the collaboration to collect the experimental data used in this paper as part of the funded Swedish Research Council project 2020-04253 \emph{Tensor-field based localization}. We also acknowledge the ELLIIT project Visionen 2.0 for partially funding of the experimental facilities used for the data collection.
\end{acks}

\bibliographystyle{SageH}
\bibliography{references.bib}


\end{document}